\documentclass[11pt,a4paper]{article}
\usepackage{emnlp2018}
\usepackage{times}
\usepackage{latexsym}
\usepackage{url}
\usepackage{enumerate}
\usepackage{graphicx}
\usepackage{caption}
\usepackage{booktabs}
\aclfinalcopy 
\newcommand{\customtilde}{\raise.17ex\hbox{$\scriptstyle\sim$}}

\title{Scaling Neural Machine Translation}

\author{%
Myle Ott$^\bigtriangleup$
\quad Sergey Edunov$^\bigtriangleup$
\quad David Grangier\hspace{1pt}$^\bigtriangledown$$^{*}$
\quad Michael Auli$^\bigtriangleup$
\\
$^\bigtriangleup$Facebook AI Research, Menlo Park \& New York. \quad \\
$^\bigtriangledown$Google Brain, Mountain View.
}

\date{}

\begin{document}
\maketitle

{\let\thefootnote\relax\footnotetext{*Work done while at Facebook AI Research.}}

\begin{abstract}
Sequence to sequence learning models still require several days to reach state of the art performance on large benchmark datasets using a single machine. This paper 
shows that reduced precision and large batch training can speedup training by nearly 5x on a single 8-GPU machine with careful tuning and implementation.\footnote{Our implementation is available at:\\ 
\url{https://www.github.com/pytorch/fairseq}}
On WMT'14 English-German translation, we match the accuracy of \citet{vaswani2017transformer} in under 5 hours when training on 8 GPUs and we obtain a new state of the art of 29.3 BLEU after training for 85 minutes on 128 GPUs.
We further improve these results to 29.8 BLEU by training on the much larger Paracrawl dataset.
On the WMT'14 English-French task, we obtain a state-of-the-art BLEU of 43.2 in 8.5 hours on 128 GPUs.

\end{abstract}

\section{Introduction}

Neural Machine Translation (NMT) has seen impressive progress in the recent years with the introduction of ever more efficient architectures~\citep{bahdanau2015neural,gehring2017convs2s,vaswani2017transformer}. Similar sequence-to-sequence models are also applied to other natural language processing tasks, such as abstractive summarization~\citep{see2017acl,paulus2018iclr} and dialog~\citep{sordoni2015acl,serban2017AHL,dusek2016seq2seq}.

Currently, training state-of-the-art models on large datasets is computationally intensive and can require several days on a machine with 8 high-end graphics processing units (GPUs). Scaling training to multiple machines enables faster experimental turn-around but also introduces new challenges:
How do we maintain efficiency in a distributed setup when some batches process faster than others (i.e., in the presence of \emph{stragglers})?
How do larger batch sizes affect optimization and generalization performance?
While stragglers primarily affect multi-machine training, questions about the effectiveness of large batch training are relevant even for users of commodity hardware on a single machine, especially as such hardware continues to improve, enabling bigger models and batch sizes.

In this paper, we first explore approaches to improve training efficiency on a single machine.
By training with reduced floating point precision we decrease training time by 65\% with no effect on accuracy.
Next, we assess the effect of dramatically increasing the batch size from 25k to over 400k tokens, a necessary condition for large scale parallelization with synchronous training.
We implement this on a single machine by accumulating gradients from several batches before each update.
We find that by training with large batches and by increasing the learning rate we can further reduce training time by 40\% on a single machine.
Finally, we parallelize training across 16 machines and find that we can reduce training time by an additional 90\% compared to a single machine.

Our improvements enable training a Transformer model on the WMT'16 En-De dataset to the same accuracy as \citet{vaswani2017transformer} in just 32 minutes on 128 GPUs and in under 5 hours on 8 GPUs.
This same model trained to full convergence achieves a new state of the art of 29.3 BLEU in 85 minutes.
These scalability improvements additionally enable us to train models on much larger datasets.
We show that we can reach 29.8 BLEU on the same test set in less than 10 hours when trained on a combined corpus of WMT and Paracrawl data containing $\sim$150M sentence pairs (i.e., over 30x more training data).
Similarly, on the WMT'14 En-Fr task we obtain a state of the art BLEU of 43.2 in 8.5 hours on 128 GPUs.

\begin{figure*}
\centering
\includegraphics[width=.45\linewidth]{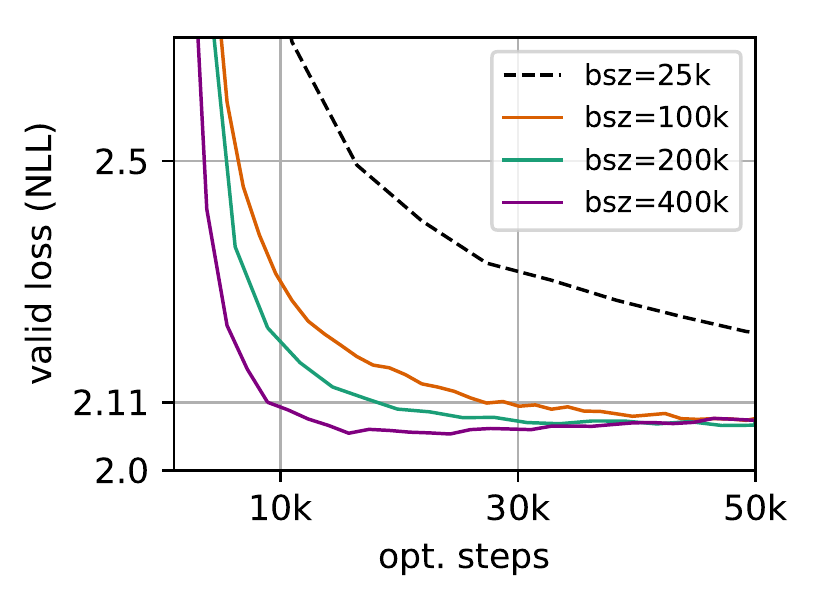}
\quad
\includegraphics[width=.45\linewidth]{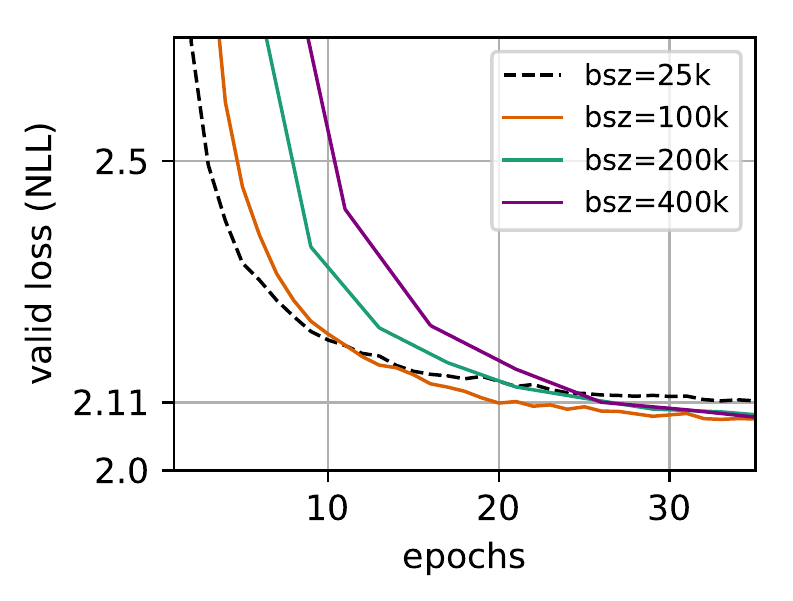}
\quad
\caption{Validation loss for Transformer model trained with varying batch sizes (bsz) as a function of optimization steps (left) and epochs (right). Training with large batches is less data-efficient, but can be parallelized. Batch sizes given in number of target tokens excluding padding. \emph{WMT En-De, newstest13}.}
\label{fig:loss_vs_steps_epochs}
\end{figure*}

\section{Related Work}

Previous research considered training and inference with reduced
numerical precision for neural networks~\cite{simard1993nips,courbariaux2015iclr,sa2018lowprec}.
Our work relies on half-precision floating point computation, following the guidelines of~\citet{narang2018iclr} to adjust the scale of the loss to avoid underflow or overflow errors in gradient computations.

Distributed training of neural networks follows two main strategies:
(i) \emph{model parallel} evaluates different model layers on different workers~\citep{coates2013hpc} and (ii) \emph{data parallel} keeps a copy of the model on each worker but distributes different batches to different machines~\citep{dean2012distributed}.
We rely on the second scheme and follow synchronous SGD, which has recently been deemed more efficient than asynchronous SGD~\citep{chen2016distributed}. Synchronous SGD distributes the computation of gradients over multiple machines and then performs a synchronized update of the model weights. Large neural machine translation systems have been recently trained with this algorithm with success~\cite{dean2017nipsw,chen2018arxiv}.

Recent work by~\citet{puri2018large} considers large-scale distributed training of language models (LM) achieving 109x scaling with 128 GPUs.
Compared to NMT training, however, LM training does not face the same challenges of variable batch sizes.
Moreover, we find that large batch training requires warming up the learning rate, whereas their work begins training with a large learning rate.
There has also been recent work on using lower precision for inference only \cite{quinn2018bit}.

Another line of work explores strategies for improving communication efficiency in distributed synchronous training setting by abandoning ``stragglers," in particular by introducing redundancy in how the data is distributed across workers~\citep{tandon2017icml,min2018icml}.
The idea rests on coding schemes that introduce this redundancy and enable for some workers to simply not return an answer.
In contrast, we do not discard any computation done by workers.

\section{Experimental Setup}

\subsection{Datasets and Evaluation}\label{sec:datasets}

We run experiments on two language pairs, English to German (En--De) and English to French (En--Fr). 
For En--De we replicate the setup of~\citet{vaswani2017transformer} which relies on the WMT'16 training data with 4.5M sentence pairs; we validate on newstest13 and test on newstest14.
We use a vocabulary of 32K symbols based on a joint source and target byte pair encoding (BPE; \citealt{bpe}).
For En--Fr, we train on WMT'14 and borrow the setup of~\citet{gehring2017convs2s} with 36M training sentence pairs.
We use newstest12+13 for validation and newstest14 for test.
The 40K vocabulary is based on a joint source and target BPE factorization.

We also experiment with scaling training beyond 36M sentence pairs by using data from the Paracrawl corpus~\cite{paracrawl}.
This dataset is extremely large and noisy with more than 4.5B pairs for En--De and more than 4.2B pairs for En--Fr.
Accordingly, we explore approaches for filtering this dataset in Section~\ref{sec:paracrawl}.
We also reuse the BPE vocabulary built on WMT data for each Paracrawl language pair.
We measure case-sensitive tokenized BLEU with \texttt{multi-bleu.pl}\footnote{\url{https://github.com/moses-smt/mosesdecoder/blob/master/scripts/generic/multi-bleu.perl}} and de-tokenized BLEU with SacreBLEU\footnote{SacreBLEU hash: \texttt{\scriptsize
\mbox{BLEU+case.mixed+lang.en-\{de,fr\}+}\\
\mbox{numrefs.1+smooth.exp+test.wmt14/full+tok.13a+}\\
\mbox{version.1.2.9}}}~\citep{post2018sacrebleu}.
All results use beam search with a beam width of 4 and length penalty of 0.6, following~\citealt{vaswani2017transformer}.
Checkpoint averaging is not used, except where specified otherwise.

\subsection{Models and Hyperparameters}

We use the Transformer model~\citep{vaswani2017transformer} implemented in PyTorch in the \texttt{fairseq-py} toolkit~\citep{fairseq}.
All experiments are based on the ``big" transformer model with 6 blocks in the encoder and decoder networks.
Each encoder block contains a self-attention layer, followed by two fully connected feed-forward layers with a ReLU non-linearity between them.
Each decoder block contains self-attention, followed by encoder-decoder attention, followed by two fully connected feed-forward layers with a ReLU between them.
We include residual connections~\cite{he2015deep} after each attention layer and after the combined feed-forward layers,
and apply layer normalization~\cite{ba2016layer} after each residual connection.
We use word representations of size 1024,
feed-forward layers with inner dimension 4,096,
and multi-headed attention with 16 attention heads.
We apply dropout~\cite{srivastava2014dropout} with probability 0.3 for En-De and 0.1 for En-Fr.
In total this model has 210M parameters for the En-De dataset and 222M parameters for the En-Fr dataset.

Models are optimized with Adam \citep{kingma2015adam} using $\beta_1 = 0.9$, $\beta_2 = 0.98$, and $\epsilon = 1\mathrm{e}{-8}$.
We use the same learning rate schedule as \citet{vaswani2017transformer}, i.e., the learning rate increases linearly for 4,000 steps to $5\mathrm{e}{-4}$ (or $1\mathrm{e}{-3}$ in experiments that specify \texttt{2x lr}),
after which it is decayed proportionally to the inverse square root of the number of steps.
We use label smoothing with $0.1$ weight for the uniform prior distribution over the vocabulary  \citep{szegedy2015inception,pereyra2017regularize}. 

All experiments are run on DGX-1 nodes with 8 NVIDIA\textsuperscript{\textcopyright} V100 GPUs interconnected by Infiniband. We use the NCCL2 library and \texttt{torch.distributed} for inter-GPU communication.

\section{Experiments and Results}\label{sec:results}

In this section we present results for improving training efficiency via reduced precision floating point (Section~\ref{sec:exp_fp16}), training with larger batches (Section~\ref{sec:large_batches}), and training with multiple nodes in a distributed setting (Section~\ref{sec:parallel}).

\begin{table*}
\centering
\begin{tabular}{l | r r r | r r r r r r}
\toprule
model & \# gpu &
\multicolumn{1}{c}{bsz} 
 & {\small cumul} & BLEU & updates & tkn/sec & time & speedup \\
\hline
\citet{vaswani2017transformer} & {\small 8$\times$P100} & 25k  & 1 &  26.4  & 300k & \customtilde 25k & \customtilde 5,000 & --\\
\hline
Our reimplementation & {\small 8$\times$V100} & 25k  & 1 &  26.4  & 192k & 54k & 1,429 & {\small reference}\\
\hline
+ \texttt{16-bit} &8& 25k  & 1 &  26.7  & 193k & 143k & 495 & 2.9x \\
+ \texttt{cumul} &8& 402k  & 16 &  26.7 &  13.7k & 195k & 447 & 3.2x\\
+ \texttt{2x lr}  &8& 402k  & 16  &  26.5 & 9.6k & 196k & 311 & 4.6x \\
+ \texttt{5k tkn/gpu}  &8& 365k & 10 & 26.5 & 10.3k & 202k & 294 & 4.9x \\
\hline
16 nodes (from {\tt +2x\,lr}) & 128 & 402k & 1 & 26.5 & 9.5k & 1.53M & 37 & 38.6x \\
+ \texttt{overlap comm+bwd} & 128 & 402k & 1 & 26.5 & 9.7k & 1.82M & 32 & 44.7x \\
\bottomrule
\end{tabular}
\caption{Training time (min) for reduced precision (\texttt{16-bit}), cumulating gradients over multiple backwards (\texttt{cumul}), increasing learning rate (\texttt{2x lr})
and computing each forward/backward with more data due to memory savings (\texttt{5k tkn/gpu}).
Average time (excl.~validation and saving models) over 3 random seeds to reach validation perplexity of $4.32$ ($2.11$ NLL).
Cumul=16 means a weight update after accumulating gradients for 16 backward computations, simulating training on 16 nodes.
\emph{WMT En-De, newstest13}.
}
\label{tab:speed}
\end{table*}

\subsection{Half-Precision Training}\label{sec:exp_fp16}

NVIDIA Volta GPUs introduce Tensor Cores that enable efficient half precision floating point (FP) computations that are several times faster than full precision operations. 
However, half precision drastically reduces the range of floating point values that can be represented which can lead to numerical underflows and overflows~\citep{narang2018iclr}.
This can be mitigated by scaling values to fit into the FP16 range.

In particular, we perform all forward-backward computations as well as the all-reduce (gradient synchronization) between workers in FP16. 
In contrast, the model weights are also available in full precision, and we compute the loss and optimization (e.g., momentum, weight updates) in FP32 as well.
We scale the loss right after the forward pass to fit into the FP16 range and perform the backward pass as usual. 
After the all-reduce of the FP16 version of the gradients with respect to the weights we convert the gradients into FP32 and restore the original scale of the values before updating the weights.

In the beginning stages of training, the loss needs to be scaled down to avoid numerical overflow, while at the end of training, when the loss is small, we need to scale it up in order to avoid numerical underflow.
Dynamic loss scaling takes care of both. 
It automatically scales down the loss when overflow is detected and since it is not possible to detect underflow, it scales the loss up if no overflows have been detected over the past 2,000 updates.

To evaluate training with lower precision, we first compare a baseline transformer model trained on 8 GPUs with 32-bit floating point (Our reimplementation) to the same model trained with 16-bit floating point (\texttt{16-bit}). Note, that we keep the batch size and other parameters equal. 
Table~\ref{tab:speed} reports training speed of various setups to reach validation perplexity 4.32 and shows that \texttt{16-bit} results in a 2.9x speedup.

\subsection{Training with Larger Batches}\label{sec:large_batches}

\begin{figure}
\begin{center}
\includegraphics[width=0.95\linewidth]{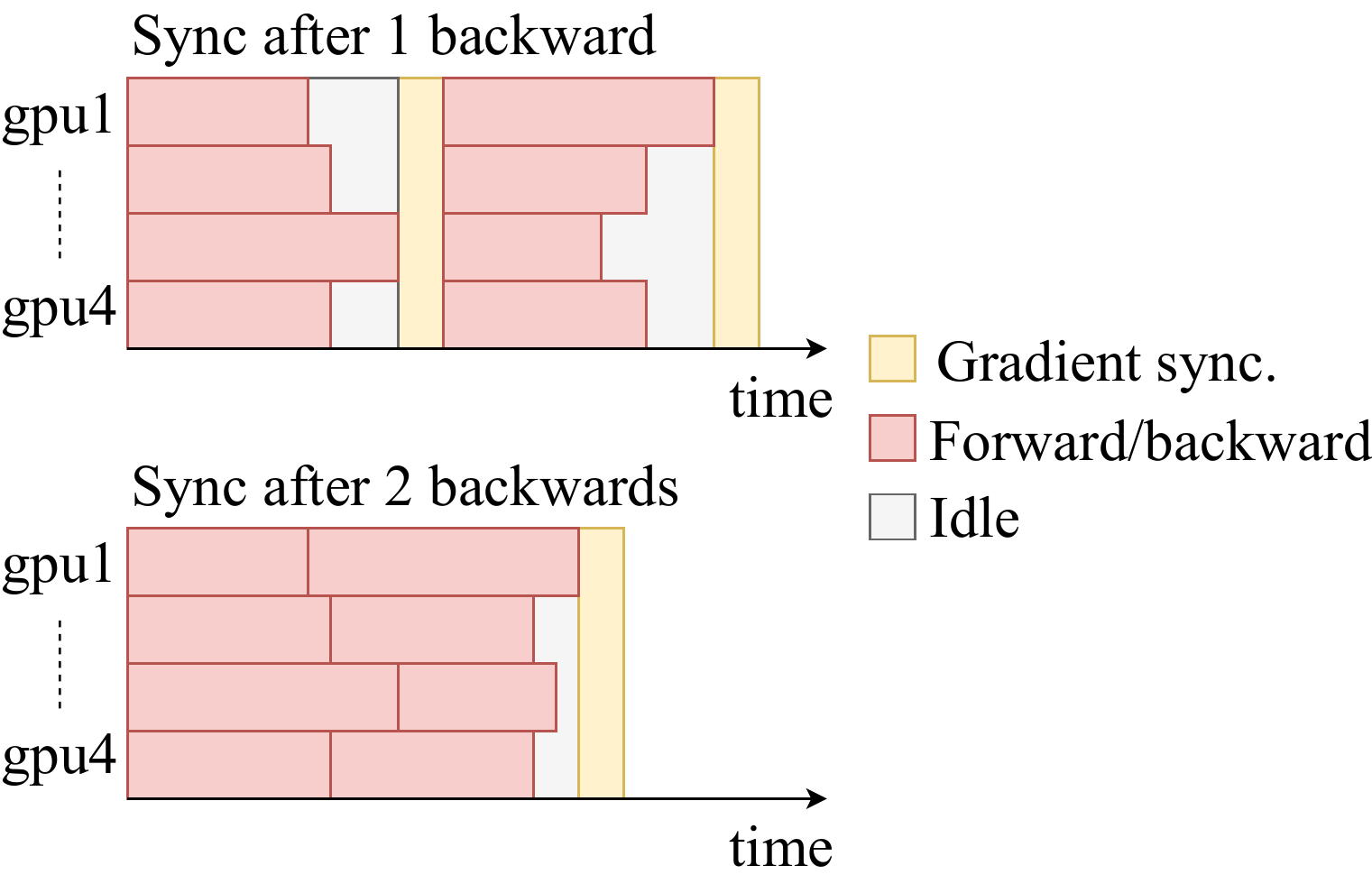}
\end{center}
\caption{Accumulating gradients over multiple forward/backward steps speeds up training by: (i) reducing communication between workers, and (ii) saving idle time by reducing variance in workload between GPUs.}
\label{fig:cumul}
\end{figure}

Large batches are a prerequisite for distributed synchronous training, since it averages the gradients over all workers and thus the effective batch size is the sum of the sizes of all batches seen by the workers.

Figure~\ref{fig:loss_vs_steps_epochs} shows that bigger batches result in slower initial convergence when measured in terms of epochs (i.e. passes over the training set). However, when looking at the number of weight updates (i.e. optimization steps) large batches converge faster~\citep{hoffer2017train}. These results support parallelization since the number of steps define the number of synchronization points for synchronous training.

Training with large batches is also possible on a single machine regardless of the number of GPUs or amount of available memory; one simply iterates over multiple batches and accumulates the resulting gradients before committing a weight update.
This has the added benefit of reducing communication and reducing the variance in workload between different workers (see Figure~\ref{fig:cumul}), leading to a 36\% increase in tokens/sec (Table~\ref{tab:speed}, \texttt{cumul}).
We discuss the issue of workload variance in more depth in Section~\ref{sec:analysis}.

\textbf{Increased Learning Rate}: Similar to~\citet{goyal2017cvpr} and \citet{smith2017lrbsz} we find that training with large batches enables us to increase the learning rate, which further shortens training time even on a single node (\texttt{2x lr}).

\textbf{Memory Efficiency}: Reduced precision also decreases
memory consumption, allowing for larger sub-batches per GPU.
We switch from a maximum of 3.5k tokens per GPU to a maximum of 5k tokens per GPU and obtain an additional 5\% speedup (cf. Table~\ref{tab:speed}; \texttt{2x lr} vs.~\texttt{5k tkn/gpu}).

Table~\ref{tab:speed} reports our speed improvements due to reduced precision, larger batches, learning rate increase and increased per-worker batch size.
Overall, we reduce training time from $1,429$ min to $294$ min to reach the same perplexity on the same hardware (8x NVIDIA V100), i.e. a 4.9x speedup.

\subsection{Parallel Training}\label{sec:parallel}

While large batch training improves training time even on a single node,
another benefit of training with large batches is that it is easily parallelized across multiple nodes (machines).
We run our previous 1-node experiment over 16 nodes of 8 GPUs each (NVIDIA V100), interconnected by Infiniband.
Table~\ref{tab:speed} shows that with a simple, synchronous parallelization strategy over 16 nodes we can further reduce training time from 311 minutes to just 37 minutes (cf. Table~\ref{tab:speed}; \texttt{2x lr} vs.~\texttt{16 nodes}).

\begin{figure}
\begin{center}
\includegraphics[width=0.95\linewidth]{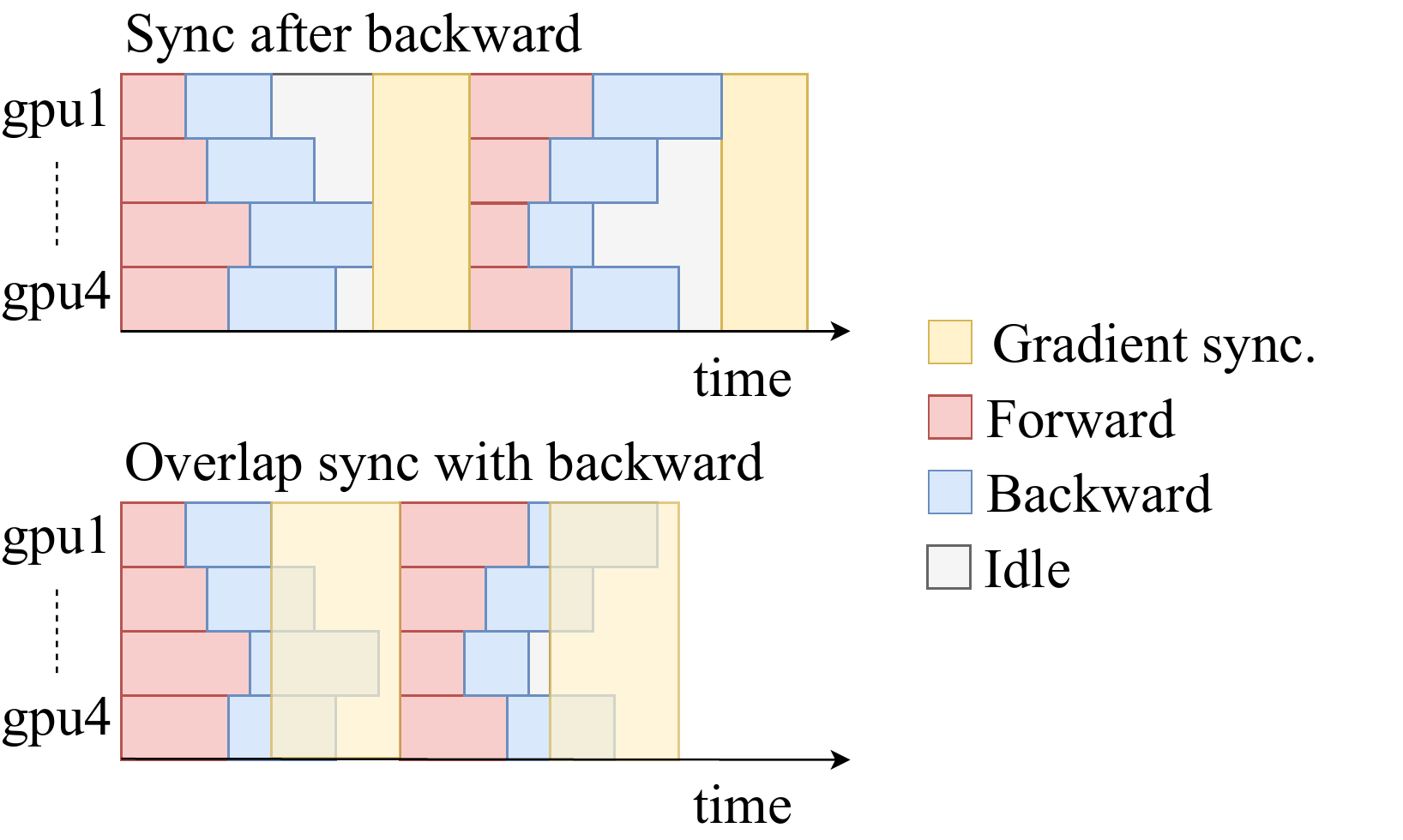}
\end{center}
\caption{Illustration of how the backward pass in back-propagation  can be overlapped with gradient synchronization to improve training speed.}
\label{fig:overlap}
\end{figure}

However, the time spent communicating gradients across workers increases dramatically when training with multiple nodes.
In particular, our models contain over 200M parameters, therefore multi-node training requires transferring 400MB gradient buffers between machines.
Fortunately, the sequential nature of back-propagation allows us to further improve multi-node training performance by beginning this communication in the background, while gradients are still being computed for the mini-batch (see Figure~\ref{fig:overlap}).
Back-propagation proceeds sequentially from the top of the network down to the inputs.
When the gradient computation for a layer finishes, we add the result to a synchronization buffer. 
As soon as the size of the buffer reaches a predefined threshold\footnote{We use a threshold of 150MB in this work.} we synchronize the buffered gradients in a background thread that runs concurrently with back-propagation down the rest of the network.
Table~\ref{tab:speed} shows that by overlapping gradient communication with computation in the backwards pass,
we can further reduce training time by 15\%, from 37 minutes to just 32 minutes (cf. Table~\ref{tab:speed}; \texttt{16 nodes} vs.~\texttt{overlap comm+bwd}).

\begin{figure}
\centering
\includegraphics[width=.95\linewidth]{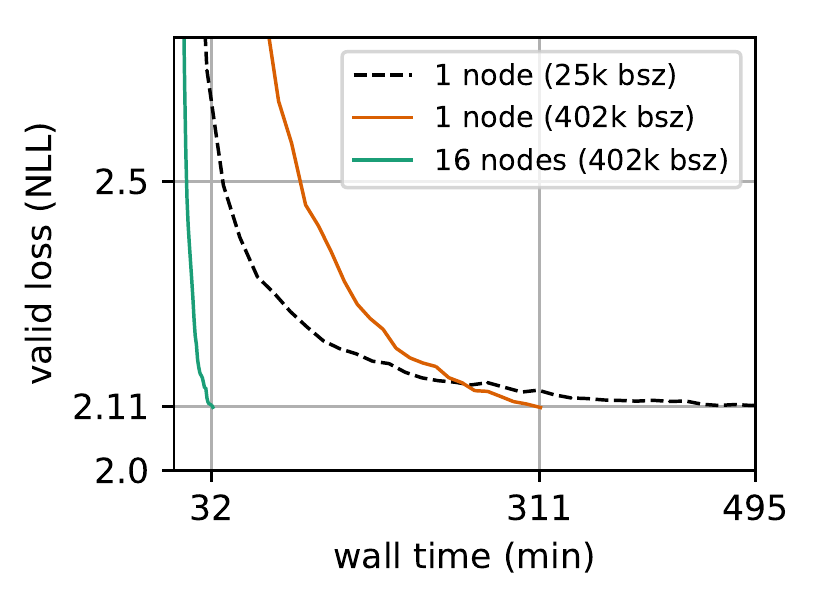}
\caption{Validation loss (negative log likelihood on newstest13) versus training time on 1 vs 16 nodes.}
\label{fig:walltime}
\end{figure}

We illustrate the speedup achieved by large batches and parallel training in Figure~\ref{fig:walltime}.

\subsection{Results with WMT Training Data}

\begin{table}[t]
\centering
\begin{tabular}{l|r r}
\toprule
                               & En--De   & En--Fr \\
\hline
a. \citet{gehring2017convs2s}    &  25.2    & 40.5 \\
b. \citet{vaswani2017transformer}&  28.4    & 41.0 \\
c. \citet{ahmed2017WeightedTN}   &  28.9    & 41.4 \\
d. \citet{shaw2018relpos}        &  29.2    & 41.5 \\
\midrule
Our result                    & {\bf 29.3}    & {\bf 43.2} \\
{\it 16-node training time}   & {\it 85 min} & {\it 512 min}\\
\bottomrule
\end{tabular}
\caption{BLEU on newstest2014 for WMT English-German (En--De) and English-French (En--Fr). All results are based on WMT'14 training data, except for En--De (b), (c), (d) and our result which are trained on WMT'16. 
}
\label{tab:testwmt}
\end{table}

\begin{table}[t]
\centering
\begin{tabular}{l|r r}
\toprule
Train set                   & En--De   & En--Fr \\
\hline
WMT only                    &  29.3  & {\bf 43.2} \\
~~~{\it detok. SacreBLEU}   	& {\it 28.6}     & {\it 41.4} \\
~~~{\it 16-node training time} & {\it 85 min} & {\it 512 min}\\
\midrule
WMT + Paracrawl             &  {\bf 29.8}  &  42.1 \\
~~~{\it detok. SacreBLEU}      &  {\it 29.3}  & {\it 40.9} \\
~~~{\it 16-node training time} & {\it 539 min} & {\it 794 min}\\
\bottomrule
\end{tabular}
\caption{Test BLEU (\emph{newstest14}) when training with WMT+Paracrawl data.\label{tab:testpara}}
\end{table}

We report results on newstest14 for English-to-German (En-De) and English-to-French (En-Fr).
For En-De, we train on the filtered version of WMT'16 from~\citet{vaswani2017transformer}.
For En-Fr, we follow the setup of~\citet{gehring2017convs2s}.
In both cases, we train a ``big'' transformer on 16 nodes and average model parameters from the last 10 checkpoints~\citep{vaswani2017transformer}.
Table~\ref{tab:testwmt} reports 29.3 BLEU for En-De in 1h 25min and 43.2 BLEU for En-Fr in 8h 32min. We therefore establish a new state-of-the-art for both datasets, excluding settings with additional training data~\citep{deepl}. In contrast to Table~\ref{tab:speed}, Table~\ref{tab:testwmt} reports times to convergence, not times to a specific validation likelihood.

\subsection{Results with WMT \& Paracrawl Training}
\label{sec:paracrawl}

Fast parallel training lets us additionally explore training over larger datasets. In this section we consider Paracrawl~\citep{paracrawl}, a recent dataset of more than 4B parallel sentences for each language pair (En-De and En-Fr).

Previous work on Paracrawl considered training only on filtered subsets of less than 30M pairs~\cite{xu2017xipporah}. 
We also filter Paracrawl by removing sentence-pairs with a source/target length ratio exceeding 1.5 and sentences with more than 250 words. We also remove pairs for which the source and target are copies~\cite{ott:uncertainty:2018}. On En--De, this brings the set from 4.6B to 700M. We then train a En--De model on a clean dataset (WMT'14 news commentary) to score the remaining 700M sentence pairs, and retain the 140M pairs with best average token log-likelihood.
To train an En--Fr model, we filter the data to 129M pairs using the same procedure.

\begin{figure}[t]
\centering
\includegraphics[width=.95\linewidth]{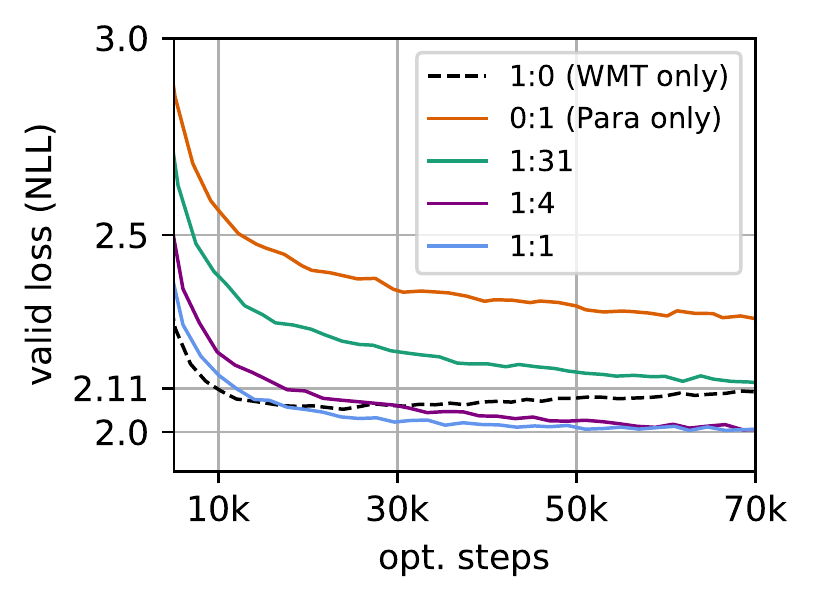}
\caption{Validation loss when training on Paracrawl+WMT with varying sampling ratios.
1:4 means sampling 4 Paracrawl sentences for every WMT sentence.
\emph{WMT En-De, newstest13}.}
\label{fig:paracrawl}
\end{figure}

Next, we explored different ways to weight the WMT and Paracrawl data.
Figure~\ref{fig:paracrawl} shows the validation loss for En-De models trained with different sampling ratios of WMT and filtered Paracrawl data during training.
The model with 1:1 ratio performs best on the validation set, outperforming the model trained on only WMT data.
For En-Fr, we found a sampling ratio of 3:1 (WMT:Paracrawl) performed best.

Test set results are given in Table~\ref{tab:testpara}.
We find that Paracrawl improves BLEU on En--De to 29.8 but it is not beneficial for En--Fr, achieving just 42.1 vs.~43.2 BLEU for our baseline.

\section{Analysis of Stragglers}\label{sec:analysis}

\begin{figure}[t]
\begin{center}
\includegraphics[width=0.95\linewidth]{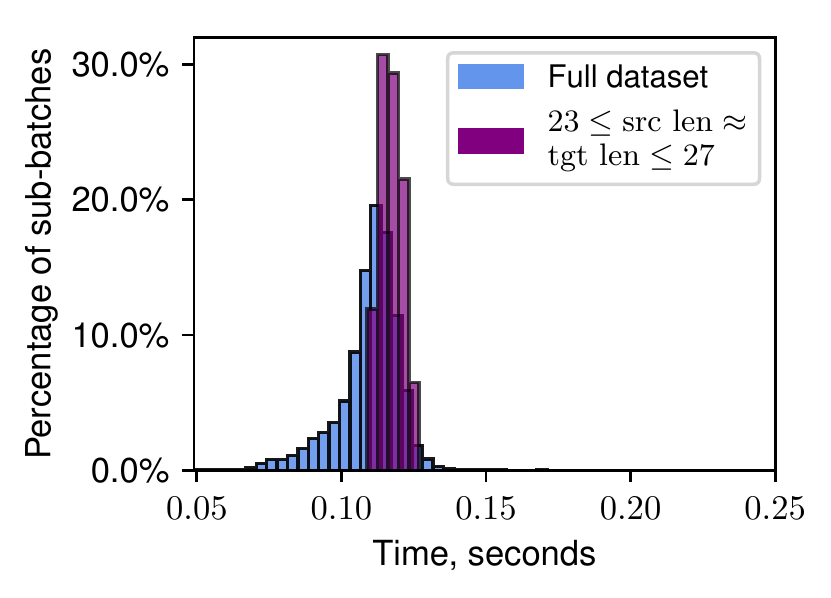}
\end{center}
\caption{
Histogram of time to complete one forward and backward pass for each sub-batch in the \emph{WMT En-De} training dataset.
Sub-batches consist of a variable number of sentences of similar length, such that each sub-batch contains at most 3.5k tokens.
}
\label{fig:batch_timing_same_vs_normal}
\end{figure}


In a distributed training setup with synchronized SGD, workers may take different amounts of time to compute gradients. 
Slower workers, or stragglers, cause other workers to wait. 
There are several reasons for stragglers but here we focus on the different amounts of time it takes to process the data on each GPU.

In particular, each GPU typically processes one \emph{sub-batch} containing sentences of similar lengths, such that each sub-batch has at most $N$ tokens (e.g., $N =$ 3.5k tokens), with padding added as required. We refer to sub-batches as the data that is processed on each GPU worker whose combination is the entire batch.
The sub-batches processed by a worker may therefore differ from other workers in the following three characteristics: the number of sentences, the maximum source sentence length, or the maximum target sentence length.
To illustrate how these characteristics impact training speed, Figure~\ref{fig:batch_timing_same_vs_normal} shows the amount of time required to process the 44K sub-batches in the En-De training data.
There is large variability in the amount time to process sub-batches with different characteristics: the mean time to process a sub-batch is 0.11 seconds, the slowest sub-batch takes 0.228 seconds and the fastest 0.049 seconds.
Notably, there is much less variability if we only consider batches of a similar shape (e.g., batches where $23 \le \textrm{src len} \approx \textrm{tgt len} \le 27$).

Unsurprisingly, constructing sub-batches based on a maximum token budget as just described exacerbates the impact of stragglers.
In Section~\ref{sec:large_batches} we observed that we could reduce the variance between workers by accumulating the gradients over multiple sub-batches on each worker before updating the weights (see illustration in Figure~\ref{fig:cumul}).
A more direct, but na\"ive solution is to assign all workers sub-batches with a similar shape.
However, this increases the variance of the gradients across batches and adversely affects the final model.
Indeed, when we trained a model in this way, then it failed to converge to the target validation perplexity of 4.32 (cf. Table~\ref{tab:speed}).

As an alternative, we construct sub-batches so that each one takes approximately the same amount of processing time across all workers.
We first set a target for the amount of time a sub-batch should take to process (e.g., the 90th percentile in Figure~\ref{fig:batch_timing_same_vs_normal}) which we keep fixed across training.
Next, we build a table to estimate the processing time for a sub-batch based on the number of sentences and maximum source and target sentence lengths.
Finally, we construct each worker's sub-batches by tuning the number of sentences until the estimated processing time reaches our target.
This approach improves single-node throughput from 143k tokens-per-second to 150k tokens-per-second, reducing the training time to reach 4.32 perplexity from 495 to 479 minutes (cf. Table~\ref{tab:speed}, \texttt{16-bit}).
Unfortunately, this is less effective than training with large batches, by accumulating gradients from multiple sub-batches on each worker (cf. Table~\ref{tab:speed}, \texttt{cumul}, 447 minutes).
Moreover, large batches additionally enable increasing the learning rate, which further improves training time (cf. Table~\ref{tab:speed}, \texttt{2x lr}, 311 minutes).

\section{Conclusions}

We explored how to train state-of-the-art NMT models on large scale parallel hardware. We investigated lower precision computation, very large batch sizes (up to 400k tokens), and larger learning rates. Our careful implementation speeds up the training of a big transformer model~\citep{vaswani2017transformer} by nearly 5x on one machine with 8 GPUs. 

We improve the state-of-the-art for WMT'14 En-Fr to 43.2 vs. 41.5 for \citet{shaw2018relpos}, training in less than 9 hours on 128 GPUs. On WMT'14 En-De test set, we report 29.3 BLEU vs. 29.2 for~\citet{shaw2018relpos} on the same setup, training our model in 85 minutes on 128 GPUs.
BLEU is further improved to 29.8 by scaling the training set with Paracrawl data.

Overall, our work shows that future hardware will enable training times for large NMT systems that are comparable to phrase-based systems~\cite{koehn:moses:2007}. We note that multi-node parallelization still incurs a significant overhead: 16-node training is only $\sim$10x faster than 1-node training. Future work may consider better batching and communication strategies.

\bibliography{main}
\bibliographystyle{acl_natbib_nourl}

\end{document}